\documentclass{article}
\usepackage{spconf,amsmath,graphicx,hyperref}
\usepackage{url}
\usepackage{cite}
\usepackage{amsmath,amssymb,amsfonts}
\usepackage{algorithm}
\usepackage{algpseudocode}
\usepackage{graphicx}
\usepackage{textcomp}
\usepackage{xcolor}
\usepackage{booktabs}
\usepackage{multirow}
\usepackage{cite}
\usepackage{amsmath,amssymb}
\usepackage{graphicx}
\usepackage{booktabs}

\title{GazeFormer-MoE: Context-Aware Gaze Estimation via CLIP and MoE Transformer}
\name{Xinyuan~Zhao$^{1}$,Xianrui~Chen$^{1}$, Ahmad~Chaddad$^{1,2,*}$ \thanks{This research was funded by the National Natural Science Foundation of China grant number 82260360, and the Guangxi Science and Technology Base and Talent Project (2022AC18004, 2022AC21040).}}
\address{\parbox{0.99 \linewidth}{\centering $^1$AIPM, School of Artificial Intelligence, Guilin University of Electronic Technology, China\\
$^2$Laboratory for Imagery, Vision and Artificial Intelligence, École de Technologie Supérieure, Canada\\ *Correspondence: ahmad8chaddad@gmail.com
}}
\begin{document}
%

\maketitle

\begin{abstract}
We present a semantics modulated, multi scale Transformer for 3D gaze estimation. Our model conditions CLIP global features with learnable prototype banks (illumination, head pose, background, direction), fuses these prototype‑enriched global vectors with CLIP patch tokens and high‑resolution CNN tokens in a unified attention space, and replaces several FFN blocks with routed/shared Mixture of Experts to increase conditional capacity. Evaluated on MPIIFaceGaze, EYEDIAP, Gaze360 and ETH‑XGaze, our model achieves new state of the art angular errors of 2.49°, 3.22°, 10.16°, and 1.44°, demonstrating up to a 64\% relative improvement over previously reported results. ablations attribute gains to prototype conditioning, cross scale fusion, MoE and hyperparameter. Our code is publicly available at \url{https://github.com/AIPMLab/Gazeformer}.
\end{abstract}
\section{Introduction}

Gaze estimation is an important task in the field of human-computer interaction (HCI), virtual and augmented reality (VR/AR), and learning analytics. It involves recovering a 3D line of sight or a 2D point of regard from images, enabling non-intrusive tracking of attention \cite{chhimpa2024revolutionizing,jayalakshmi2024multi,li2024gaze,mathew2024gescam}. Recent advances in deep learning have taken advantage of full-face input to benefit from explicit head pose signals \cite{cheng2024appearance,zhang2017s}, although this can introduce variability due to person-specific morphology, background clutter, and varying illumination, which harms robustness to appearance variation and cross‑condition performance \cite{liu2024pnp}. CLIP offers alignment of semantic priors that can be integrated without additional annotations \cite{radford2021learning}. Sparse Mixture-of-Experts (MoE) modules enhance conditional capacity efficiently by engaging a selective set of experts for each token \cite{fedus2022switch,riquelme2021scaling}. However, current frameworks frequently lack the capability for targeted, sample-wise semantic modulation or fail to synthesize global semantics with mid- and fine-grained tokens, coupled with adaptive expert routing \cite{cheng2024appearance}. In response, we introduce GazeFormer-MoE, a streamlined framework that conditions CLIP global features via learned prototype banks, merges these enriched global vectors with CLIP patch and high-resolution CNN tokens within a unified Transformer model, and applies a routed/shared MoE for diversified appearance handling. This innovative design improves robustness against changes in illumination, pose, and background, while preserving interpretability. Our contributions are summarized as follows.

\begin{enumerate}
      \item \textbf{Method.} We propose a semantics‑modulated multi scale pipeline that injects CLIP aligned, learnable prototypes into global features and jointly attends prototype‑enriched global vectors, CLIP patch tokens and high‑resolution CNN tokens within a single Transformer encoder.
    \item \textbf{Architecture.} We design a routed and shared MoE Transformer that combines specialized experts for rare appearance sub distributions with shared experts for stability, increasing modeling capacity without uniformly increasing dense parameters.
    \item \textbf{Evaluation.} We adhere to the benchmark defined in a recent review of gaze estimation \cite{cheng2024appearance}, and based on this, attains new state of the art (SOTA) results on four existing benchmarks. 
\end{enumerate}
The remainder of this paper is organized as follows. Section~\ref{methodology} formalizes the gaze estimation task and details our approach, including CLIP prototype selection, unified multi scale token fusion, and routed/shared MoE Transformer design. Section~\ref{experiment} presents datasets, implementation details, and training protocols. Section~\ref{discussion} provides a comprehensive discussion, and Section~\ref{conclusion} summarizes the study.

\begin{figure*}[h]
     \centering
\includegraphics[width=1\linewidth]{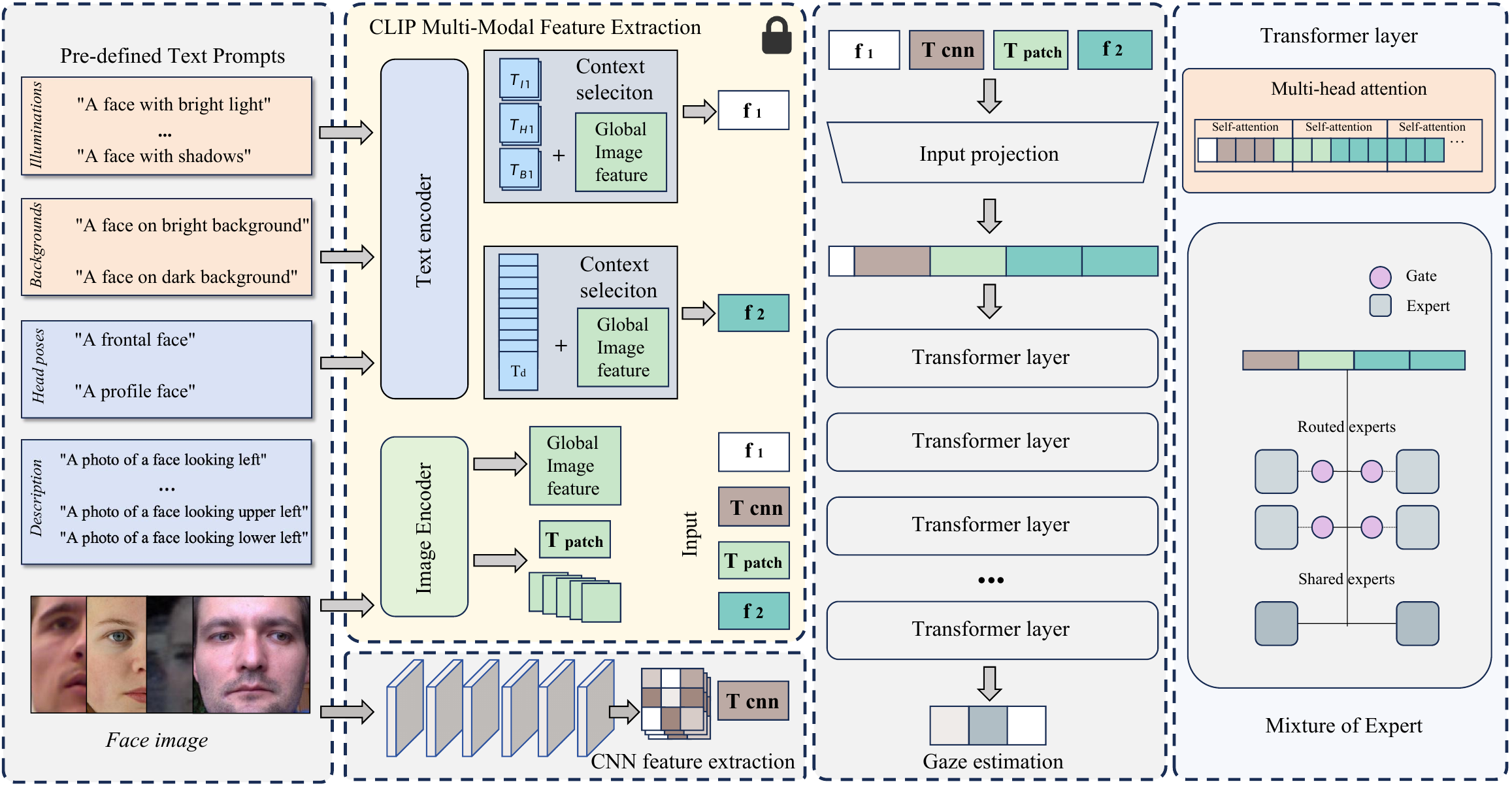}
\vspace{-1.7em}
\caption{\textbf{(Left) CLIP Feature Extraction:} A pre-trained CLIP text encoder processes four prompt categories (Illuminations, Backgrounds, Head Poses, Descriptions), which are fused with global image features via context selection to form semantic features $\mathbf{f}_1,\mathbf{f}_2$. Meanwhile, a CNN extracts local visual maps.
\textbf{(Right) Transformer Gaze Estimation:} Multi-modal features ($\mathbf{f}1$, $\mathbf{f}2$, global patches $\mathbf{T}_{\text{patch}}$, CNN features $\mathbf{T}_{\text{cnn}}$) are projected, concatenated, and input into a Transformer encoder with MoE layers. The gating-based MoE enhances capacity and adaptability, yielding the final gaze prediction.}
    \label{pipeline}
\end{figure*}

\section{Methodology}
\label{methodology}
We present GazeFormer-MoE, a multi-modal Transformer that combines CLIP-driven semantic context, patch-level tokens and high-resolution CNN features to estimate 3D gaze directions. The detailed process is summarized in Algorithm~\ref{alg:pipeline_simplified}
\begin{algorithm}[ht]
\caption{GazeFormer-MoE}
\label{alg:pipeline_simplified}
\begin{algorithmic}[1]
\Require Image $I$, ground truth gaze $\mathbf{g}$
\Ensure Predicted gaze $\hat{\mathbf{g}}$

\State \Comment{1. Extract multi-scale features from image $I$}
\State $\mathbf{f}_{\mathrm{global}}, \mathbf{T}_{\mathrm{patch}} \gets \mathcal{E}_i(I)$ \Comment{CLIP features}
\State $\mathbf{T}_{\mathrm{cnn}} \gets \mathrm{Flatten}(\mathcal{F}_{\text{cnn}}(I))$ \Comment{CNN features}

\Statex
\State \Comment{2. Condition global feature with semantic prototypes}
\State For each context $c$, select best prototype $\mathbf{P}_c[p_c]$ via similarity (Eq. \ref{eq:context-sel})
\State Form two enriched vectors $\mathbf{f}_1, \mathbf{f}_2$ using the selected prototypes (Eq. \ref{eq:context-vectors})

\Statex
\State \Comment{3. Fuse tokens and process with MoE-Transformer}
\State $\mathcal{X} \gets \text{Concatenate}(\mathbf{f}_1, \mathbf{f}_2, \mathbf{T}_{\text{patch}}, \mathbf{T}_{\text{cnn}})$ \Comment{Create sequence (Eq. \ref{eq:seq})}
\State $\mathcal{Y} \gets \text{MoE-Transformer}(\mathcal{X})$ \Comment{Apply MoE blocks (Eq. \ref{eq:moe})}

\Statex
\State \Comment{4. Predict gaze and compute loss}
\State $\hat{\mathbf{g}} \gets \text{PredictionHead}(\text{Pool}(\mathcal{Y}))$
\State $\mathcal{L}_{\text{total}} \gets \mathcal{L}_{\text{ang}}(\hat{\mathbf{g}},\mathbf{g}) + \text{regularizers}$ \Comment{Using angular loss (Eq. \ref{eq:ang-loss})}

\State \Return $\hat{\mathbf{g}}$
\end{algorithmic}
\end{algorithm}
\subsection{Problem formulation}

Given a face image \(I \in \mathbb{R}^{H \times W \times 3}\) of height \(H\) and width \(W\), the task is to predict a 3D unit gaze vector \(\mathbf{g} \in \mathbb{R}^3\). We denote by \(\mathcal{E}_i\) a pre-trained CLIP vision encoder and by \(\mathcal{F}_{\text{cnn}}\) a trainable CNN backbone. From \(I\), three types of features are extracted:

\begin{equation}
\begin{aligned}
\mathbf{f}_{\mathrm{global}} &= \mathcal{E}_i(I)\in\mathbb{R}^{d},\\
\mathbf{T}_{\mathrm{patch}} &= \{\mathbf{t}_1,\dots,\mathbf{t}_N\}\in\mathbb{R}^{N\times d},\\
\mathbf{T}_{\mathrm{cnn}} &= \mathrm{Flatten}(\mathbf{F}_{\mathrm{cnn}})\in\mathbb{R}^{M\times C},
\end{aligned}
\end{equation}

\noindent where \(\mathbf{f}_{\mathrm{global}}\) is the CLIP global embedding of dimension \(d\), \(\mathbf{T}_{\mathrm{patch}}\) are \(N\) CLIP patch tokens, and \(\mathbf{T}_{\mathrm{cnn}}\) are \(M=H'W'\) tokens obtained by flattening the CNN feature map \(\mathbf{F}_{\mathrm{cnn}} \in \mathbb{R}^{H'\times W'\times C}\) with spatial size \(H'\!\times\! W'\) and channel dimension \(C\).

\subsection{Context-aware semantic conditioning}
We maintain learnable prototype banks \(\mathbf{P}_c\) for contexts \(c\in\{\text{illum},\text{head},\text{bg},\text{label}\}\). For each image, similarity scores $s_c$ between the normalized global feature and prototypes are computed:
\begin{equation}\label{eq:context-sel}
s_c = \text{softmax}\big(\tau\cdot (\hat{\mathbf{f}}_{\text{global}})(\hat{\mathbf{P}}_c)^\top\big),
\end{equation}
where \(\tau\) is a learnable temperature. We select the top prototype \(p_c=\arg\max s_c\) and form two context-conditioned vectors:
\begin{equation}\label{eq:context-vectors}
\begin{aligned}
\mathbf{f}_1 &= \mathrm{LayerNorm}\Big(\mathbf{f}_{\mathrm{global}}
  + \sum_{c\in\{\mathrm{illum,head,bg}\}}\mathbf{P}_c[p_c]\Big),\\[4pt]
\mathbf{f}_2 &= \mathrm{LayerNorm}\Big(\mathbf{f}_{\mathrm{global}}
  + \mathbf{P}_{\mathrm{label}}[p_{\mathrm{label}}]\Big).
\end{aligned}
\end{equation}
This supplies sample-wise semantic priors without extra annotations.

\subsection{Unified multi-scale token fusion}
We concatenate the enriched global vectors with token sequences to form the Transformer input $\mathcal{X}$:
\begin{equation}\label{eq:seq}
\mathcal{X} = \big[\mathbf{f}_1;\ \mathbf{f}_2;\ \mathbf{T}_{\text{patch}};\ \mathbf{T}_{\text{cnn}}\big],
\end{equation}

\subsection{Mixture of Experts Transformer}\label{eq:moe}
Our Transformer replaces feed-forward sublayers with gated MoE blocks. For token \(\mathcal{X}\), routing produces sparse expert weights \(r(\mathcal{X})\) and the MoE output is:
\begin{equation}\label{eq:moe}
\text{MoE}(\mathcal{X})=\sum_{i\in\text{TopK}(r(\mathcal{X}))} r_i(\mathcal{X})\cdot \text{Expert}_i(\mathcal{X}).
\end{equation}
We use a routed and shared design: a set of specialized experts (route dependent) plus shared experts for base capacity. Top-K, expert count \(E\) (refer to section \ref{implementationdetails}), and the load-balancing loss follow standard practice.

\subsection{Prediction head and training loss}
The Transformer output is pooled and mapped to a 3D vector \(\hat{\mathbf{g}}\). We train with an angular objective that is numerically stable:
\begin{equation}\label{eq:ang-loss}
\mathcal{L}_{\text{ang}}(\hat{\mathbf{g}},\mathbf{g}) = 1 - \frac{\hat{\mathbf{g}}^\top \mathbf{g}}{\|\hat{\mathbf{g}}\|\|\mathbf{g}\|},
\end{equation}
which is monotone with the angular error and avoids explicit arccos computation. We add standard regularizers (weight decay, MoE load balancing) and optimize with AdamW.

\vspace{1em}
\begin{table}[h]
\centering
\caption{Comparison of GE error (in degrees) on multiple datasets. Lower is better.}
\label{tab:gaze_comparison}
\setlength{\tabcolsep}{0.5pt}
\begin{tabular}{lccccc}
\toprule
\textbf{Methods} & \textbf{Pub. Year} & \textbf{\textbf{M}} & \textbf{\textbf{E}} & \textbf{\textbf{G}} & \textbf{\textbf{Et}} \\
\midrule

Gazenet \cite{zemblys2019gazenet}          & TPAMI17     & 5.76$^\circ$  & 6.79$^\circ$  &-& -        \\
FullFace \cite{zhang2017s}          & CVPR17     & 4.93$^\circ$  & 6.53$^\circ$  &14.99$^\circ$& 7.38$^\circ$        \\
Dilated-Net \cite{chen2018appearance}          & ACCV19     & 4.42$^\circ$  & 6.19$^\circ$  &13.73$^\circ$& -        \\
Gaze360 \cite{kellnhofer2019gaze360}          & ICCV19     & 4.06$^\circ$  & 5.36$^\circ$  &11.04$^\circ$& 11.04$^\circ$        \\
CA-Net \cite{gu2020net}          & AAAI 20     & 4.27$^\circ$  & 5.27$^\circ$  & 11.20$^\circ$ & -        \\
AFF-Net \cite{du2024aff}                 & ICPR 20     & 4.92$^\circ$  & 6.41$^\circ$  & -         & -        \\
GazeTR-Hybrid \cite{cheng2022gaze}          & ICPR 22     & 4.18$^\circ$  & 5.44$^\circ$  & 11.46$^\circ$ & -        \\
GazeTr-Pure \cite{9956687}          & ICPR 22     & 4.24$^\circ$  & 5.72$^\circ$  &13.58$^\circ$& -        \\
GazeCLIP \cite{wang2023gazeclip}       & arXiv 25    & 3.50$^\circ$   & 4.70$^\circ$   & -         & -        \\
CLIP-DFENet \cite{zhang2025clip}    & arXiv 25    & 3.71$^\circ$  & 4.97$^\circ$  & 10.54$^\circ$ & -        \\
MCA-PGI \cite{li2025nonlinear}    & S. Reports 25    & 3.90$^\circ$  & 4.58$^\circ$  & 10.34$^\circ$ & -        \\
GazeSymCAT \cite{zhong2025gazesymcat}    & JCDE 25    & 4.11$^\circ$  & 5.13$^\circ$  & - & 3.28$^\circ$        \\
IGTGGaze \cite{nie2025iris}    & TIP 25    & 3.60$^\circ$  & 4.56$^\circ$  & 10.92$^\circ$ & -        \\
PCNet \cite{tian2025disengage}    & TIP 25    & 3.99$^\circ$  & 4.50$^\circ$  & - & 4.00$^\circ$        \\
\textbf{Ours} & -          & \textbf{2.49$^\circ$} & \textbf{3.22$^\circ$} & \textbf{10.16$^\circ$} & \textbf{1.44$^\circ$}        \\

\bottomrule
\end{tabular}
\end{table}
\vspace{-1.5em}


\section{Experiment}
\label{experiment}
\subsection{Implementation Details}
\label{implementationdetails}
Experiments are conducted on NVIDIA RTX 4090 with PyTorch 2.4.1+cu124. The model combines CLIP ViT-B/32, ResNet-50, and a 12-layer MoE Transformer (8 heads, 512 dim, 2048 FFN), Top 4 experts. The MoE has 8 experts (4 active, 4 shared, 1024 FFN), with a unified feature dimension of 768. Training uses AdamW (LR $10^{-4} \rightarrow 10^{-6}$, cosine annealing), 100 epochs, batch size 128.
\subsection{Datasets}

We evaluate GazeFormer-MoE on four widely used GE benchmarks: EYEDIAP (\textbf{E})~\cite{funesmora2014eyediap}, ETH-XGaze (\textbf{Et})~\cite{zhang2020ethxgaze}, Gaze360 (\textbf{G})~\cite{kellnhofer2019gaze360}, and MPIIFaceGaze (\textbf{M})~\cite{zhang2017fullface}. These datasets jointly span controlled indoor setups, wide headpose and appearance variation, and in-the-wild conditions, enabling evaluation under diverse illumination, background, and pose regimes. Following the benchmark protocol of \cite{cheng2024appearance}, we strictly divide the dataset into training validation and testing set with a ratio of 8 : 1 : 1. 

\subsection{Experimental results}
Our method establishes new state of the art performance across all four benchmarks (Table~\ref{tab:gaze_comparison}). On \textbf{M} and \textbf{E} we reduce the best previously reported errors (3.5$^\circ$ / 4.50$^\circ$) to 2.49$^\circ$ / 3.22$^\circ$. \textbf{G} remains the most challenging; we still improve the strongest prior result (10.34$^\circ$) to 10.16$^\circ$ while preserving robustness at large yaw. The largest absolute and relative gain appears on \textbf{E} (4.00$^\circ$ $\rightarrow$ 1.44$^\circ$), consistent with our design goal of handling illumination and background diversity: unified multi-scale token fusion plus routed and shared MoE markedly mitigates overfitting to studio-like lighting and captures fine periocular shading cues.

\subsubsection{Ablation study}
Table~\ref{tab:ablation1} shows that prototype-conditioned global vectors alone yield high errors, adding high-resolution CNN tokens produces the largest error collapse, and CLIP patch tokens provide the final consistent refinement. Table~\ref{tab:ablation2} isolates the effect of routed/shared MoE: removing MoE degrades performance across all benchmarks (most markedly on \textbf{E} and \textbf{M}), confirming that sparse conditional experts improve performance. Table~\ref{tab:ablation3} reports learning‑rate sensitivity and identifies $10^{-4}$ as the best choice for stable convergence; both larger and smaller rates hurt accuracy.
\label{ablations}

\vspace{-1.5em}
\begin{table}[h]
    \centering\small
    \setlength{\tabcolsep}{4.5pt}
    \renewcommand{\arraystretch}{0.9}  
    \caption{Experimental Results of Ablation Studies on Different Feature Combinations‌}
    \label{tab:ablation1}
    \begin{tabular}{lcccc}
        \toprule
        \multirow{2}{*}{\textbf{Features combination}} & \multicolumn{4}{c}{\textbf{Datasets}} \\
        \cmidrule(lr){2-5}
        & \textbf{M} & \textbf{E} & \textbf{G} & \textbf{Et} \\
        \midrule
        $\mathbf{f}_1$ & 7.66$^\circ$ & 10.25$^\circ$ & 28.43$^\circ$ & 10.75$^\circ$ \\
        $\mathbf{f}_1 + \mathbf{f}_2$ & 7.72$^\circ$ & 10.15$^\circ$ & 27.70$^\circ$ & 10.40$^\circ$ \\
        $\mathbf{f}_1 + \mathbf{f}_2 + \mathbf{T}_{\text{cnn}}$ & 3.20$^\circ$ & 4.39$^\circ$ & 10.92$^\circ$ & 1.66$^\circ$\\
        $\mathbf{f}_1 + \mathbf{f}_2 + \mathbf{T}_{\text{cnn}} + \mathbf{T}_{\text{patch}}$ & 2.49$^\circ$ & 3.22$^\circ$ & 10.16$^\circ$ & 1.44$^\circ$        \\
        \bottomrule
    \end{tabular}
\end{table}
\vspace{-1.5em}
\begin{table}[ht]
    \centering\small
    \setlength{\tabcolsep}{6pt}
    \renewcommand{\arraystretch}{0.9}
    \caption{Experimental Results of Ablation Studies on MoE Effect‌}
    \label{tab:ablation2}
    \begin{tabular}{lcccc}
        \toprule
        \multirow{2}{*}{\textbf{MoE}} & \multicolumn{4}{c}{\textbf{Datasets}} \\
        \cmidrule(lr){2-5}
        & \textbf{M} & \textbf{E} & \textbf{G} & \textbf{Et} \\
        \midrule
        $+$ & 2.49$^\circ$ & 3.22$^\circ$ & 10.16$^\circ$ & 1.44$^\circ$ \\
        w/o &4.20$^\circ$ &5.78$^\circ$&10.72$^\circ$ &4.38$^\circ$ \\
        \bottomrule
    \end{tabular}
\end{table}
\vspace{-1.5em}
\begin{table}[ht]
    \centering\small
    \setlength{\tabcolsep}{8pt}
    \renewcommand{\arraystretch}{0.8}
    \caption{Experimental Results of Ablation Studies on Different Learning rate}
    \label{tab:ablation3}
    \begin{tabular}{lcccc}
        \toprule
        \multirow{2}{*}{\textbf{Learning rate}} & \multicolumn{4}{c}{\textbf{Datasets}} \\
        \cmidrule(lr){2-5}
        & \textbf{M} & \textbf{E} & \textbf{G} & \textbf{Et} \\
        \midrule
        $10^{-3}$ & 8.46$^\circ$ & 11.30$^\circ$ & 10.90$^\circ$ & 4.66$^\circ$ \\
        $10^{-4}$& 2.49$^\circ$ & 3.22$^\circ$ & 10.16$^\circ$ & 1.44$^\circ$ \\
        $10^{-5}$ & 3.74$^\circ$ & 7.70$^\circ$ & 15.61$^\circ$ & 1.84$^\circ$\\
       $10^{-6}$& 5.39$^\circ$ & 9.81$^\circ$ & 15.89$^\circ$ & 3.70$^\circ$ \\
        $10^{-7}$& 5.79$^\circ$ & 10.37$^\circ$ & 17.38$^\circ$ & 4.23$^\circ$ \\
        \bottomrule
    \end{tabular}
\end{table}
\vspace{-1.5em}
\section{Discussion}
\label{discussion}
The evidence shows that coupling CLIP-aligned prototype conditioning with unified multi-scale token fusion and a routed/shared MoE materially reshapes the generalization profile of appearance-based gaze estimation under illumination, pose, and background shift. Achieving 2.49$^\circ$, 3.22$^\circ$, 10.16$^\circ$, and 1.44$^\circ$ on datasets \textbf{M}, \textbf{E}, \textbf{G}, and \textbf{Et} respectively (up to 64\% relative improvement over the strongest prior reports) . Ablation studies clarify that the gains are not a simple depth or parameter scaling effect: prototype-conditioned global tokens alone remain weakly constrained (errors above 7$^\circ$, 10$^\circ$, 28$^\circ$, and 10$^\circ$ on the four datasets), indicating that coarse semantic priors without fine spatial structure under-express critical periocular micro-texture and shading cues; introducing high-resolution CNN tokens collapses much of this gap (e.g., \textbf{M} 7.66$^\circ$ to 3.20$^\circ$), and the subsequent addition of CLIP patch tokens delivers a further consistent refinement (\textbf{M} 3.20$^\circ$ to 2.49$^\circ$; \textbf{E} 4.39$^\circ$ to 3.22$^\circ$), supporting the hypothesis that low-level texture, mid-level semantic structure, and prototype-guided context are complementary when co-attended within a single sequence space rather than fused late. Removing MoE produces uniform regressions (e.g., \textbf{Et} 1.44$^\circ$ to 4.38$^\circ$), suggesting that a single dense feed-forward path simultaneously underfits rare, illumination-perturbed or high-yaw sub-distributions and overfits dominant frontal well-lit modes. The discrete prototype selection (argmax with learnable temperature) helps sharpen semantic routing and avoid noisy soft mixtures but imposes a finite granularity that may under-represent continuous illumination gradients or composite mixed-spectrum indoor–outdoor transitions; a static vocabulary also risks amplifying CLIP domain biases under sensor spectral shift (e.g., low-light color cast or infrared leakage). MoE introduces practical trade-offs: routing variance and potential tail latency under unbalanced expert loads, plus sensitivity to the load-balancing coefficient which, if mis-tuned, reduces capacity exactly where extreme yaw or deep shadow samples would benefit. Further, the present formulation is strictly single-frame and does not exploit temporal coherence (micro-saccades, blink dynamics, head micro-motion) that could regularize transient noise. 
\vspace{-1.0em}
\section{Conclusion}
\label{conclusion}
We presented GazeFormer-MoE, a semantics-modulated multi-scale Transformer that combines CLIP-driven prototype selection, unified fusion of prototype-enriched global vectors, CLIP patch tokens, and CNN tokens, and a routed and shared MoE to adapt capacity across different modalities. The model achieves state of the art errors of 2.49$^\circ$, 3.22$^\circ$, 10.16$^\circ$, and 1.44$^\circ$ on datasets \textbf{M}, \textbf{E}, \textbf{G}, and \textbf{Et}, delivering up to 64\% relative improvement. Ablations show (i) semantic prototypes alone are insufficient, (ii) cross-scale token unification supplies complementary texture and mid-level structure, and (iii) MoE routing is critical for handling long-tail (shadowed, extreme yaw) cases without overfitting dominant regimes. The approach remains limited by a static and discrete prototype vocabulary, single-frame processing, and routing overhead. Future directions include dynamic prototype evolution, temporally aware sparse experts, and latency-oriented expert distillation. Overall, results support semantic conditioning plus conditional capacity as a concise recipe for robust fine-grained geometric estimation.
\section{Compliance with Ethical Standards}
\vspace{-9pt}

This is a numerical simulation study for which no ethical approval was required.
\footnotesize
\bibliographystyle{ieeetr}

\bibliography{mybib}

@article{cheng2024appearance,
  title={Appearance-based gaze estimation with deep learning: A review and benchmark},
  author={Cheng, Yihua and Wang, Haofei and Bao, Yiwei and Lu, Feng},
  journal={IEEE Transactions on Pattern Analysis and Machine Intelligence},
  volume={46},
  number={12},
  pages={7509--7528},
  year={2024},
  publisher={IEEE}
}

@article{chhimpa2024revolutionizing,
  title={Revolutionizing Gaze-based Human-Computer Interaction using Iris Tracking: A Webcam-Based Low-Cost Approach with Calibration, Regression and Real-Time Re-calibration},
  author={Chhimpa, Govind Ram and Kumar, Ajay and Garhwal, Sunita and Khan, Faheem and Moon, Yeon-Kug and others},
  journal={IEEE Access},
  year={2024},
  publisher={IEEE}
}

@inproceedings{jayalakshmi2024multi,
  title={Multi-model Human-Computer Interaction System with Hand Gesture and Eye Gesture Control},
  author={Jayalakshmi, M and Saradhi, T Pardha and Azam, Syed Mohammed Rahil and Fazil, Sk and Sriram, S Durga Sai},
  booktitle={2024 5th International Conference on Innovative Trends in Information Technology (ICITIIT)},
  pages={1--6},
  year={2024},
  organization={IEEE}
}

@article{li2024gaze,
  title={E-gaze: Gaze estimation with event camera},
  author={Li, Nealson and Chang, Muya and Raychowdhury, Arijit},
  journal={IEEE Transactions on Pattern Analysis and Machine Intelligence},
  volume={46},
  number={7},
  pages={4796--4811},
  year={2024},
  publisher={IEEE}
}

@inproceedings{mathew2024gescam,
  title={GESCAM: A Dataset and Method on Gaze Estimation for Classroom Attention Measurement},
  author={Mathew, Athul M and Khan, Arshad Ali and Khalid, Thariq and Souissi, Riad},
  booktitle={Proceedings of the IEEE/CVF Conference on Computer Vision and Pattern Recognition},
  pages={636--645},
  year={2024}
}

@inproceedings{zhang2017s,
  title={It's written all over your face: Full-face appearance-based gaze estimation},
  author={Zhang, Xucong and Sugano, Yusuke and Fritz, Mario and Bulling, Andreas},
  booktitle={Proceedings of the IEEE conference on computer vision and pattern recognition workshops},
  pages={51--60},
  year={2017}
}

@article{liu2024pnp,
  title={Pnp-ga+: Plug-and-play domain adaptation for gaze estimation using model variants},
  author={Liu, Ruicong and Liu, Yunfei and Wang, Haofei and Lu, Feng},
  journal={IEEE Transactions on Pattern Analysis and Machine Intelligence},
  volume={46},
  number={5},
  pages={3707--3721},
  year={2024},
  publisher={IEEE}
}

@inproceedings{funesmora2014eyediap,
  author    = {Kenneth Alberto Funes Mora and Florent Monay and Jean-Marc Odobez},
  title     = {{EYEDIAP}: A database for the development and evaluation of gaze estimation algorithms from {RGB} and {RGB-D} cameras},
  booktitle = {Proceedings of the ACM Symposium on Eye Tracking Research and Applications (ETRA)},
  year      = {2014},
  pages     = {255--258},
  publisher = {ACM}
}

@inproceedings{zhang2020ethxgaze,
  author    = {Xucong Zhang and Seonwook Park and Thabo Beeler and Derek Bradley and Siyu Tang and Otmar Hilliges},
  title     = {{ETH-XGaze}: A large scale dataset for gaze estimation under extreme head pose and gaze variation},
  booktitle = {Proceedings of the European Conference on Computer Vision (ECCV)},
  year      = {2020},
  pages     = {365--381},
  publisher = {Springer}
}

@inproceedings{kellnhofer2019gaze360,
  author    = {Petr Kellnhofer and Adria Recasens and Simon Stent and Wojciech Matusik and Antonio Torralba},
  title     = {{Gaze360}: Physically unconstrained gaze estimation in the wild},
  booktitle = {Proceedings of the IEEE International Conference on Computer Vision (ICCV)},
  year      = {2019},
  pages     = {2176--2184},
  publisher = {IEEE}
}

@inproceedings{zhang2017fullface,
  author    = {Xucong Zhang and Yusuke Sugano and Mario Fritz and Andreas Bulling},
  title     = {It's written all over your face: Full-face appearance-based gaze estimation},
  booktitle = {Proceedings of the IEEE Conference on Computer Vision and Pattern Recognition Workshops (CVPRW)},
  year      = {2017},
  pages     = {2299--2308},
  publisher = {IEEE}
}

@article{fedus2022switch,
  title={Switch Transformers: Scaling to Trillion Parameter Models with Simple and Efficient Sparsity},
  author={Fedus, William and Zoph, Barret and Shazeer, Noam},
  journal={JMLR},
  year={2022}
}

@inproceedings{riquelme2021scaling,
  title={Scaling Vision with Sparse Mixture of Experts},
  author={Riquelme, Carlos and Puigcerver, Joan and Mustafa, Basil and et al.},
  booktitle={NeurIPS},
  year={2021}
}

@INPROCEEDINGS{9956687,
  author={Cheng, Yihua and Lu, Feng},
  booktitle={2022 26th International Conference on Pattern Recognition (ICPR)}, 
  title={Gaze Estimation using Transformer}, 
  year={2022},
  volume={},
  number={},
  pages={3341-3347},
  doi={10.1109/ICPR56361.2022.9956687}}

@inproceedings{chen2018appearance,
  title={Appearance-based gaze estimation using dilated-convolutions},
  author={Chen, Zhaokang and Shi, Bertram E},
  booktitle={Asian Conference on Computer Vision},
  pages={309--324},
  year={2018},
  organization={Springer}
}

@article{zemblys2019gazenet,
  title={gazeNet: End-to-end eye-movement event detection with deep neural networks},
  author={Zemblys, Raimondas and Niehorster, Diederick C and Holmqvist, Kenneth},
  journal={Behavior research methods},
  volume={51},
  number={2},
  pages={840--864},
  year={2019},
  publisher={Springer}
}

@article{gu2020net,
  title={CA-Net: Comprehensive attention convolutional neural networks for explainable medical image segmentation},
  author={Gu, Ran and Wang, Guotai and Song, Tao and Huang, Rui and Aertsen, Michael and Deprest, Jan and Ourselin, S{\'e}bastien and Vercauteren, Tom and Zhang, Shaoting},
  journal={IEEE transactions on medical imaging},
  volume={40},
  number={2},
  pages={699--711},
  year={2020},
  publisher={IEEE}
}

@article{du2024aff,
  title={AFF-Net: A strip steel surface defect detection network via adaptive focusing features},
  author={Du, Yongzhao and Chen, Haixin and Fu, Yuqing and Zhu, Jianqing and Zeng, Huanqiang},
  journal={IEEE Transactions on Instrumentation and Measurement},
  volume={73},
  pages={1--14},
  year={2024},
  publisher={IEEE}
}

@inproceedings{cheng2022gaze,
  title={Gaze estimation using transformer},
  author={Cheng, Yihua and Lu, Feng},
  booktitle={2022 26th International Conference on Pattern Recognition (ICPR)},
  pages={3341--3347},
  year={2022},
  organization={IEEE}
}

@article{wang2023gazeclip,
  title={Gazeclip: Towards enhancing gaze estimation via text guidance},
  author={Wang, Jun and Ruan, Hao and Wang, Mingjie and Zhang, Chuanghui and Li, Huachun and Zhou, Jun},
  journal={arXiv preprint arXiv:2401.00260},
  year={2023}
}

@article{zhang2025clip,
  title={CLIP-driven Dual Feature Enhancing Network for Gaze Estimation},
  author={Zhang, Lin and Tian, Yi and Xu, Wanru and Jin, Yi and Huang, Yaping},
  journal={arXiv e-prints},
  pages={arXiv--2502},
  year={2025}
}

@article{li2025nonlinear,
  title={Nonlinear multi-head cross-attention network and programmable gradient information for gaze estimation},
  author={Li, Yujie and Hong, Yuhang and Wang, Ziwen and Chen, Jiahui and Liu, Rongjie and Ding, Shuxue and Tan, Benying},
  journal={Scientific Reports},
  volume={15},
  number={1},
  pages={27135},
  year={2025},
  publisher={Nature Publishing Group UK London}
}

@article{zhong2025gazesymcat,
  title={GazeSymCAT: A symmetric cross-attention transformer for robust gaze estimation under extreme head poses and gaze variations},
  author={Zhong, Yupeng and Lee, Sang Hun},
  journal={Journal of Computational Design and Engineering},
  volume={12},
  number={3},
  pages={115--129},
  year={2025},
  publisher={Oxford University Press}
}

@article{nie2025iris,
  title={Iris Geometric Transformation Guided Deep Appearance-Based Gaze Estimation},
  author={Nie, Wei and Wang, Zhiyong and Ren, Weihong and Zhang, Hanlin and Liu, Honghai},
  journal={IEEE Transactions on Image Processing},
  year={2025},
  publisher={IEEE}
}

@article{tian2025disengage,
  title={‘Disengage AND Integrate’: Personalized Causal Network for Gaze Estimation},
  author={Tian, Yi and Wang, Xiyun and Zhang, Sihui and Xu, Wanru and Jin, Yi and Huang, Yaping},
  journal={IEEE Transactions on Image Processing},
  year={2025},
  publisher={IEEE}
}

@inproceedings{radford2021learning,
  title={Learning transferable visual models from natural language supervision},
  author={Radford, Alec and Kim, Jong Wook and Hallacy, Chris and Ramesh, Aditya and Goh, Gabriel and Agarwal, Sandhini and Sastry, Girish and Askell, Amanda and Mishkin, Pamela and Clark, Jack and others},
  booktitle={International conference on machine learning},
  pages={8748--8763},
  year={2021},
  organization={PmLR}
}

\end{document}